\documentclass[sigconf]{acmart}

\AtBeginDocument{%
  }

\setcopyright{acmcopyright}
\copyrightyear{2018}
\acmYear{2018}
\acmDOI{XXXXXXX.XXXXXXX}

\acmConference[Conference acronym 'XX]{Make sure to enter the correct
  conference title from your rights confirmation emai}{June 03--05,
  2018}{Woodstock, NY}
\acmPrice{15.00}
\acmISBN{978-1-4503-XXXX-X/18/06}

\usepackage[utf8]{inputenc} 
\usepackage[T1]{fontenc}    
\usepackage{hyperref}       
\usepackage{url}            
\usepackage{booktabs}       
\usepackage{amsfonts}       
\usepackage{nicefrac}       
\usepackage{microtype}      
\usepackage{xcolor}         
\usepackage{graphicx}
\usepackage{subcaption}
\usepackage{lineno}
\usepackage{booktabs}
\usepackage{tabularx}

\begin{document}
\graphicspath{ {images/} }

\title{OpportunityFinder: A Framework for Automated Causal Inference}

\author{Huy Nguyen}
\authornote{Both authors contributed equally to this research.}
\affiliation{%
  \institution{Amazon.com}
  \country{USA}
}
\email{nguynnq@amazon.com}

\author{Prince Grover}
\authornotemark[1]
\affiliation{%
  \institution{Amazon.com}
  \country{USA}
}
\email{pringrov@amazon.com}

\author{Devashish Khatwani}
\affiliation{%
  \institution{Amazon.com}
  \country{USA}
}
\email{khatwad@amazon.com}


\begin{abstract}
  We introduce OpportunityFinder, a code-less framework for performing a variety of causal inference studies with panel data for non-expert users. In its current state, OpportunityFinder only requires users to provide raw observational data and a configuration file. A pipeline is then triggered that inspects/processes data, chooses the suitable algorithm(s) to execute the causal study. It returns the causal impact of the treatment on the configured outcome, together with sensitivity and robustness results. Causal inference is widely studied and used to estimate the downstream impact of individual's interactions with products and features. It is common that these causal studies are performed by scientists and/or economists periodically. Business stakeholders are often bottle-necked on scientist or economist bandwidth to conduct causal studies. We offer OpportunityFinder as a solution for commonly performed causal studies with four key features: (1) easy to use for both Business Analysts and Scientists, (2) abstraction of multiple algorithms under a single I/O interface, (3) support for causal impact analysis under binary treatment with panel data and (4) dynamic selection of algorithm based on scale of data.
\end{abstract}



\maketitle

\section{Introduction}
\label{headings}
Automated machine learning (AutoML) frameworks for predictive machine learning (ML) have advanced significantly over the past decade with the introductions of AutoGluon \cite{agtabular}, Auto-sklearn \cite{feurer-arxiv20a}, H2O \cite{H2OAutoML20}. AutoML's biggest advantage is abstracting away the implementation of underlying algorithms and hyper-parameter tuning, and making it easy for scientists and engineers to experiment with a large number of models and identify the one that works best.
The demand of AutoML has risen from the fact that no single ML algorithm works best in all scenarios. This has been even more challenging in the causal inference literature. Different methods rely on different set of assumptions \cite{causalAssumptions} for the identification of causal treatment effects\footnote{A causal effect can be defined as the difference between hypothetical outcomes that result from two or more alternative treatments, with only one outcome of a treatment being observed each time}: CIA (conditional independence assumption or unconfoundedness), propensity overlap, SUTVA (stable unit treatment value assignment), exchangeability (same outcome distribution would be observed if exposed and unexposed individuals were exchanged) etc.


Causal inference framework DoWhy \cite{dowhy} supports explicit modeling and testing of causal assumptions, but it is still a low level API.
AutoCausality, \cite{autocausality} which is built on the top of EconML \cite{econml} and DoWhy, supports automated hyperparameter tuning, but it only focuses on the estimation part and assumes that the causal graph provided by the user accurately explains data-generating process. Both AutoCausality and DoWhy do not support panel data\footnote{Panel data contains observations collected across multiple individuals at a regular frequency, and ordered chronologically.}, which is a mainstream at real-world problems. 
Most real-world causal studies have panel data of different aggregated granularities, e.g., yearly to daily levels, at different scales, e.g., few individuals to large populations of million entities.
To the best of our knowledge, there is no AutoML-like causal inference framework that supports panel data and abstracts away the know-how of causal studies from the users.

In this study, we introduce OpportunityFinder (OPF), our first step in democratizing causal inference techniques. As of our first contribution, Project OPF implements an auto causal inference framework that supports panel and cross-sectional data and offers a wide range of causal inference algorithms. The decision to choose the algorithm is automated and abstracted away from the user. Our second contribution is the automated transformation from input panel data into list of cohort datasets when needed. Cohort-based results are then aggregated for a final result. In the third contribution, OPF provides data visualization to illustrate causal impact. Combining numerical and graphical reports help non-expert users to verify input data and reason about causal inference results. Current capability of OPF allows non-expert users to carry out the most common causal analysis: \emph{estimating the average treatment effect (ATE) with configurable time horizons for binary actions}. At current state, OpportunityFinder is deployed within AWS account of our organization for internal testing. We are also refactoring OpportunityFinder source as a stand-alone library.



\begin{figure*}[t]
  \centering
  \begin{minipage}[t]{0.35\textwidth}
    \vspace{0pt} 
    \begin{verbatim}
    1. ###### API: End-to-end ######
    
    2. from opportunity_finder.api \
    3. import OpportunityFinder
    
    4. opf = OpportunityFinder(
    5.         treatment_df, 
    6.         observations_df,
    7.         config_dict)
            
    8. opf.estimate_causal_effect()
    \end{verbatim}
  \end{minipage}\hfill%
  \begin{minipage}[t]{0.64\textwidth}
    \begin{minipage}[t]{0.14\textwidth}
      \vspace{0pt} 
      \begin{tabular}{cc}
        \hline
        \small{Unit ID} & \small{Treatment Date} \\
        \hline
        A & 2022-07-23 \\
        B & 2022-07-16 \\
        C & 2022-10-01 \\
        D & 2023-01-11 \\
        E & 2022-03-05 \\
        F & 2023-03-21 \\
        G & 2022-05-05 \\
        H & 2022-10-12 \\
        I & 2022-12-02 \\
        J & 2022-02-25 \\
        \hline
      \end{tabular}
    \end{minipage}\hfill%
    \begin{minipage}[t]{0.62\textwidth}
      \vspace{0pt} 
      \begin{tabularx}{\textwidth}{ccccc}
        \hline
        \small{Unit ID} & \small{Date} & \small{Impressions} & \small{Clicks} & \small{Sales} \\
        \hline
        A & 2022-07-01 & 3.43M & 400K & \$0.25M \\
        A & 2022-07-08 & 3.62M & 410K & \$0.26M \\
        A & 2022-07-15 & 3.90M & 423K & \$0.39M \\
        A & 2022-07-22 & 4.21M & 431K & \$0.32M \\
        A & 2022-07-29 & 3.52M & 399K & \$0.40M \\
        Z & 2022-07-01 & 8.12M & 912K & \$10.1M \\
        Z & 2022-07-08 & 8.42M & 923K & \$10.1M \\
        Z & 2022-07-15 & 8.55M & 922K & \$10.3M \\
        Z & 2022-07-22 & 8.21M & 942K & \$8.1M \\
        Z & 2022-07-29 & 8.12M & 890K & \$11.2M \\
        \hline
      \end{tabularx}
    \end{minipage}
  \end{minipage}
    \caption{Left: Sample of OpportunityFinder UX with Python. Center: Sample treatment data. Right: Sample observational data with possible set of covariates.}

  \label{fig:snapshots}
\end{figure*}

\section{Literature Review}
\label{lit}

Traditional econometric techniques such as propensity score matching, instrumental variable estimation, and difference-in-differences (DiD), offer rigorous methods for estimating average treatment effects under specific assumptions, but often struggle to account for high-dimensional covariates and complex interactions
\cite{Angrist2009}. The Synthetic Control Method (SCM) extends these approaches by constructing a ``synthetic'' control unit as a weighted combination of potential control units, providing a more flexible comparison for the treated unit
\cite{Abadie2003}. The Generalized Synthetic Control (GSC) further expands SCM by incorporating interactive fixed effects models, thus accommodating multiple treated units and variable treatment periods
\cite{xu_2017}.

Recently, machine learning techniques have been widely integrated into causal inference due to notable works by various teams, e.g., DoubleML~\cite{DoubleML2022Python}, EconML~\cite{econml}, CausalML~\cite{chen2020causalml}.
Double Machine Learning (DML) provides a flexible approach, leveraging machine learning for nuisance parameter estimation while maintaining robustness against mis-specification
\cite{Chernozhukov2018}. Beyond average treatment effect, machine learning enables approaches to estimate individual treatment effects, e.g., heterogeneous treatment effect estimator in EconML and uplift modeling in CausalML. Deep learning methods, such as those based on Neural Networks (NN), have shown promise in estimating individual treatment effects due to their ability to model complex, high-dimensional data, thus uncovering nuanced causal relationships \cite{Shalit2017}.


\section{Framework Design}

The key contributions of our design are (1) integration of several causal modeling models, (2) branching based on type of observational data (cross sectional vs. panel) and number of treatment units, and (3) execution in the users' own AWS environment where they have access to CloudWatch logs for debugging and can visualize the progress. Current OpportunityFinder deployment allows code-less UI without having to move data outside the AWS account as demonstrated in Figure~\ref{fig:snapshots}. While this design is tied to the MLOps set-up of our organization, OpportunityFinder source code is independent from deployment platforms.

\begin{figure*}[t]
\includegraphics[scale=0.4]{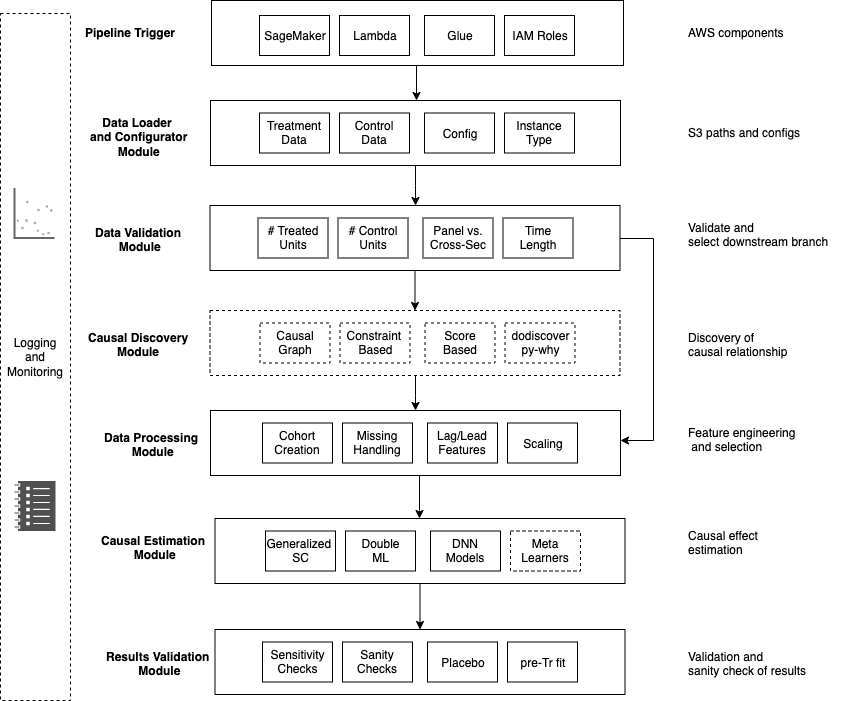}
\centering
\caption{Framework design of OpportunityFinder. \textit{Dotted boxes are under planned development.}}
\label{fig:high-level-design}
\end{figure*}

Figure~\ref{fig:high-level-design} shows the design of OpportunityFinder.
Once a user triggers a job, CloudFormation kicks off a set of AWS services including SageMaker, Lambda and Glue jobs. Data Validation module checks treatment and control data for basic requirements. SageMaker Pipeline then starts with performing follow-up components. Data Processing transforms panel data into cohorts (where needed), handles missing data, extracts lag/lead features, performs optional data scaling and normalization. Causal Estimation decides most suitable causal model given data, and executes the model. Result Validation performs validation tests for sanity and sensitivity, and returns the estimated treatment effect in a standardized format into the user's S3 bucket.

The data processing can vary for different underlying models. For example, Generalized Synthetic Control (GSC) \cite{xu_2017} works well even if there is one treated unit, but it requires panel data with at-least 7 pre-treatment periods. 
Double Machine Learning (DML) \cite{Chernozhukov2018} is a better solution for large-scale data but requires breaking down treatments into the cohorts of different weeks, months or quarters, depending on the number of treated individuals in each cohort.

On completion of causal estimation, a series of sensitivity and placebo tests are applied to assess the robustness of the findings to violations of the underlying assumptions. These validations include (but not limited to), direction of causal relationship, sensitivity of causal estimate to small variations in observations data (e.g., down-sampling, random co-variate) and variations in model hyper-parameters (e.g., number of pre-treatment periods used for finding synthetic controls). The results of these validation tests are written to the S3 bucket for user reference.

\subsection{Data Requirements}
\label{data-req}
OpportunityFinder requires user to provide two datasets and a configuration file (examples shown in Figure \ref{fig:snapshots}). The first data, i.e., treatment data, should contain IDs of the treated units \footnote{Individuals, e.g., shoppers, advertisers, who activated a feature or received a treatment} and date when the treatment happened.
The second data, also known as, baseline observational or control data, contains the observational information about all IDs that received treatment as well as the ones that did not receive treatment during the same period. 
Control data should contain time-based, e.g., daily, weekly or monthly, outcome variables (i.e., target) of interest such as ad spend, click count over the historical period. At the same level of time granularity, user is recommended to add a superset of possible variables (i.e., features) that are related to the outcome and the treatment. Among those superset of variables, the model will search for the ones that can help in removing the confounding and mediating effects, an essential for accurate causal estimates.

Configuration file has optional and mandatory requirements. Optional requirements like list of  features to scale, choice of algorithm, choice of hyper-parameters allow user flexibility, but are not necessary and can be automatically handled by the framework. The mandatory requirements include  columns that specify time, unit id, outcome variable and pre/post-treatment evaluation window, e.g., 4 weeks, 6 months. Based on user-provided configuration and data validation, input panel data might be segmented into cohorts and feature engineering would be performed, before passing to causal analysis algorithms.

\subsection{Implementation Details}

\subsubsection{Two Stage Decision Path}
\label{dtree}

The decision of causal estimation algorithm goes through two stages. First stage is a set of rules, based on factors that include the following. Depending on the answers of these factors, a causal estimation algorithm is selected (leaf node of the decision path) 

\begin{itemize}
    \item Are total event data less than or more than 500,000?
    \item Is the data panel or cross sectional?
    \item Are number of treated units per cohort less than or more than 50?
    \item Are number of control units less than or more than 5,000?
    \item Are number of covariates as per causal graph more or less than 5?
    \item How many periods (e.g., daily, monthly) of pre and post-treatment data are available?
\end{itemize}

In scenarios, where the above set of rules give more than one option of causal estimators, the decision flow moves to the second stage. It tries out the list of all models chosen in the first stage, and selects the result that has least \emph{standard error} on the estimated output and within lower and upper bounds of at-least 2 more estimators (\emph{voting mechanism}).

\subsubsection{Cohort Data}

One of key functions provided by OpportunityFinder is the transformation of panel data into cohort, i.e., sectional data, which allows techniques like double machine learning to work. Each cohort corresponds to a set of treated units that received treatment in a closed period. First, treatment data is processed to extract list of treatment times and number of treated units at each time. A cohort is a set of one, two or more consecutive treatment times and constrained by three parameters: Minimum/maximum number of treatment times, and minimum number of treated units.

For example, if treatment happens at day level, the first two parameters specify the lower and upper bounds of number of days in each cohort. The third parameter says that a cohort must have at least a certain number of treated units. We sort treatment times in ascending order, for each treatment time we keep merging it with following times until three conditions above are satisfied. Then a list of cohorts is returned. Results from each cohort are aggregated using weighted average w.r.t number of treated units in each cohort.

\subsubsection{Causal Inference Models}

OpportunityFinder implements a wide set of popular and advanced causal inference algorithms, as following. Further details of these and other models that we considered are in Appendix
\ref{app:models}.

\begin{itemize}
    \item Generalized Synthetic Controls: our implementation is based on original R code \cite{xu_2017}.
    \item Double Machine Learning: using EconML package \cite{econml} we employ two treatment effect estimators, LinearDML and CausalForestDML. Each estimator is stacked with any of four classes of base prediction models that predict treatment and outcome: Random Forest, Linear Regression, XGBoost, and LightGBM.
    \item Deep Neural Networks: we implement four state of the arts DNN algorithms for estimation of treatment effect, BCAUSS \cite{tesei2023learning}, DRAGON \cite{shi2019adapting}, TARNET \cite{Shalit2017}, and GANITE \cite{yoon2018ganite}.
\end{itemize}

\subsubsection{Validation Tests}

DML models and their treatment effect estimation are validated through refutation tests by DoWhy package: add random common cause, add unobserved common causes, data subsets validation, and placebo treatment. For a robust causal model and valid treatment effect, first three tests should return treatment effect similar to original model while fourth test must have effect close to zero. GSC model is validated with a suite of sensitivity tests that check for changes in the estimated causal effect with small changes in data like random down-sampling, different 
pre-treatment window for learning synthetic control weights and a reduced covariate list. The 
expectation is that the causal effect should not change the direction of estimation 
with small changes in the setting. Example test results are shown in Appendix~\ref{test:val}. In future, we will equip validation tests for DNN models.

\subsubsection{Data Visualization}

A challenge that prevents the adoption of causal inference studies is a lack of ground-truth data which makes estimation error impossible to assess. OpportunityFinder addresses this by providing visualizations that naively explain the treatment effect to some extent. For example, it returns a plot that shows the trend of outcome for treated and control units over time. The visualizations are a part of Logging and Monitoring module. While such plots do not confirm calculated treatment effect by causal models, they help non-expert users to comprehend causal inference results. Example visualizations are shown in Appendix~\ref{app:vis}.

\subsection{Limitations and Risks}
\label{limitations}

As of today, OpportunityFinder (OPF) does not implement causal graph generation algorithms. This also means that the tool has less flexibility for someone who wants to control covariates and experiment with different algorithms. We plan to integrate causal discovery module in near future.  

OPF applies our best heuristics after exploring input data to select the right algorithm. Due to the lack of ground truth data in causal inference, our framework can make mistakes without knowing that the estimated effect it returns is wrong. We select a model based on \emph{standard error} and ensemble by voting to mitigate this limitation upto some extent. The accuracy of estimate still depends on the observational data given by the user.

For real-world problems, OPF does not necessarily use estimation models that gave best score on benchmark data.
Our experiments show that simpler estimators work more reasonably than DNN models on our use-case data.

\section{Validation of Causal Estimates by OpportunityFinder}
\label{val}

We validate our causal inference algorithms on benchmark datasets using three metrics, choosing the metric available in related research for each dataset.





\begin{enumerate}
\item Average Treatment Effects (ATE): measures causal impact of a treatment/intervention on a population by comparing the average hypothetical outcomes between receiving and not receiving treatment, accounting for potential confounding factors. 

\item Average Treatment Effects on Treated (ATT): is ATE measured on treated units.

\item Mean Absolute Error (MAE): the average absolute difference between estimated ATE and true ATE where available, for evaluating accuracy of a causal estimation method.
\end{enumerate}



\subsection{IHDP \textit{(public benchmark)}}
\label{ihdp}

The Infant Health and Development Program (IHDP) \cite{ihdbbench} is a randomized controlled study designed to evaluate the effect of specialist visit on cognitive test scores of premature infants.
This dataset is cross-sectional data, with binary treatment (specialist visits), continuous outcome (cognitive scores) and has known ground truth ATE. As shown in Table~\ref{tab:ihdb}, our implementation of DML models achieved competitive performance. Results on BCAUSS, TARNET and DRAGON are based on our implementation and slightly differ from reported numbers \cite{ihdbbench}. The difference is because DNN methods are executed within OPF pipeline and data are not prepared as the same as previous studies.

\begin{table*}[t]
\centering
\caption{Mean absolute error on IHDP benchmark. All models are part of OpportunityFinder}
\label{tab:ihdb}
{\normalsize
\begin{tabular}{rrr|rrrr}
\toprule
\multicolumn{3}{c|}{DNN} & \multicolumn{4}{c}{+LinearDML}\\
\midrule
\footnotesize{BCAUSS} &
\footnotesize{DRAGON} &
\footnotesize{TARNET} &
\footnotesize{LinearReg.} &
\footnotesize{Rand.Forest} &
\footnotesize{XGBoost} &
\footnotesize{LightGBM} \\
\midrule
0.23 &
0.32 &
0.25 &
0.42 &
0.48 &
0.47 &
0.43 \\
\bottomrule
\end{tabular}
}
\end{table*}

\subsection{Smoking \textit{(public benchmark)}}
The goal of smoking data is to analyze the causal effect of Proposition 99 on cigarette sales. This data has small size with just one treated unit, thus causal estimations based on machine learning (DML, NN) do not apply.
OpportunityFinder chooses to run GSC and does not create any cohorts.
Table \ref{tab:smoking} shows comparison of results using OPF on this dataset with previous research \cite{doi:10.1198/jasa.2009.ap08746, arkhangelsky2021synthetic} works. We observe that the range of ATE estimate lies between -11.1 to -27.1, and the results from OPF are within the range of previous studies.\footnote{Synthetic difference in differences (SDID), Synthetic controls (SC), Difference in differences (DID), Matrix completion (MC), Synthetic control with intercept (DIFP).} 
When we use cigarette retail price as a covariate, the ATE reduces to -14.0, which is closer to SDID. It is discussed in SDID \cite{arkhangelsky2021synthetic} paper that their results (-15.6) are more credible among other approaches shown in the table. We also observe a lower standard error with OPF results. This experiment helps validate OPF on small panel dataset. 

\begin{table*}[t]
\centering
\caption{Estimates for ATE on Smoking data
}
\label{tab:smoking}
{\normalsize
\begin{tabular}{lrrrrr|rr}
\toprule
 &
SDID &
SC &
DID &
MC &
DIFP &
OPF &
OPF w.price \\
  \midrule
ATE &
  -15.6 &
  -19.6 &
  -27.3 &
  -20.2 &
  -11.1 &
  -24.6 &
  \textbf{-14.0} \\
Standard Error &
  8.4 &
  9.9 &
  17.7 &
  11.5 &
  9.5 &
  4.9 &
  \textbf{4.7} \\
  \bottomrule
\end{tabular}%
}
\end{table*}



\subsection{Synthetic Data 1: Cross Sectional \textit{(synthetic)}}

In addition to public datasetes, we validate OpportunityFinder outputs on two 
synthetic datasets with known ATE.\footnote{github.com/groverpr/datasets}
The first synthetic dataset is a linear cross sectional dataset that we generated using DoWhy \cite{dowhy} package. We created the dataset with 2 instrument, 5 common causes, 5000 samples and binary treatment with some treatment noise.
Because this data is cross sectional, OPF rejects GSC, but branches off to the second stage of decision path, where it evaluates multiple models, including DML \cite{Chernozhukov2018}, and neural net based estimators including BCAUSS \cite{tesei2023learning},  TARNET \cite{Shalit2017}, DRAGON \cite{shi2019adapting} and GANITE \cite{yoon2018ganite}.
As we see in Table \ref{tab:synthetic-data}, all models (except GANITE) give ATE close to the true ATE. OPF finally selects the model with least standard error if the mean is within the range of 2+ other models, and 
ends up selecting \texttt{LinearReg+LinearDML}. 

\begin{table*}[t]
\centering
\caption{ATE (with std. error) on synthetic datasets using models implemented in OPF}
\label{tab:synthetic-data}
{\normalsize
\begin{tabular}{lrlrl}
\toprule
 &                                   \multicolumn{2}{c}{Synthetic\#1 (GT = 10)}        & \multicolumn{2}{c}{Synthetic\#2 (GT = 20)} \\
 \midrule
 BCAUSS                            & 10.18 &(0.03)         &       19.00 &(0.18)        \\
 DRAGON                         & 10.34 &(0.19)         &       18.98 &(0.24)        \\
 TARNET                            & 10.00 &(0.09)         &       18.98 &(0.23)       \\
 GANITE                            & 7.78 &(n/a)        &        6.41 &(n/a)     \\
 LinearReg. + LinearDML & \textbf{9.93} &(0.02) &        19.04 &(0.13) \\     
 Rand.Forest + LinearDML     & 9.98 &(0.07)          &            19.01 &(0.13) \\
 XGBoost + LinearDML   & 9.70 &(0.04)          &             18.82 &(0.13) \\
 LightGBM + LinearDML & 9.75 &(0.03)          &              18.97 &(0.15) \\
 GSC                               & n/a  &                & \textbf{18.87} & (0.07)   \\
\bottomrule
\end{tabular}%
}
\end{table*}

\subsection{Synthetic Data 2: Large Panel \textit{(synthetic)}}

In the second dataset, we add non-linear confounding effect and correlated variables on a panel data, to test the efficacy of different supported models to be able to remove the bias. This data contains 52,000 rows, 3 confounders with non linear effect on treatment and outcome, 1000 units, 263 treated units and 52 time periods. The properties of this dataset enables OpportunityFinder to run all implemented algorithms and select based on standard errors.
As shown in Table \ref{tab:synthetic-data}, all models except GANITE, perform well that estimated ATE are close to ground-truth. Due to lowest variance in estimations from GSC together with mean estimation lying between 2+ other estimators, OPF chooses GSC results for the end user.

\subsection{Discussion on Model Choice}
While our approach for model selection is evolving, the results on synthetic and public datasets show that our current 2 stage decision path works well. The two stage decision path allows automated rejection of estimators if they are not built for the use case at hand. For example, GSC is supposed to be used for panel data but becomes computationally inefficient with >500,000 data sizes. OpportunityFinder does not run GSC for such large data sizes. 
As the research evolves, especially with neural networks for causal inference, we plan to incorporate the new models as well as update the models selection criteria. For example, we will explore providing results from ensemble of models
and finding the expected causal path using causal discovery algorithms.  

\section{Applications on Real World Data}
\label{prod}

OpportunityFinder has already been used in multiple use cases. In this section, we present two most important applications of OPF within our organization.
Most commonly used down stream impact metric in real applications is \emph{uplift}, which is defined as the percentage increase/decrease in the outcome attributed to the treatment over a defined period. It is calculated as ATE or ATT divided by average over control units. 

\subsection{Opportunity for Partners}


\begin{table*}[t]
\centering
\caption{A sample of opportunity for partners studies from 2021/22 vs 2023. Metric is uplift after 6 months of adoption.
}
\label{tab:hvo}
{\normalsize
\begin{tabular}{lcccccc}
\toprule
 & \multicolumn{3}{c}{Opportunity \emph{X}} & \multicolumn{3}{c}{Opportunity \emph{Y}} \\
\cmidrule(lr){2-4} \cmidrule(lr){5-7}
 & Metric 1 & Metric 2 & Metric 3 & Metric 1 & Metric 2 & Metric 3  \\
\midrule
Manual (2021/22) & 5\% & 20\% & 12\% & 4\% & 4\% & 6\% \\
OpportunityFinder (2023) & 6\% & 12\% & 17\% & 8\% & 8\% & 11\% \\
\bottomrule
\end{tabular}
}
\end{table*}

\begin{table*}[t]
\centering
\caption{Results of opportunity for advertisers study for world-wide vendors. Metric is average monthly uplift on outcomes within 3 months after adoption.}
\label{tab:vap}
{\normalsize
\begin{tabular}{lrrrrrrrr}
\toprule
& \multicolumn{4}{c}{DNN} & \multicolumn{4}{c}{+LinearDML}\\
\cmidrule(lr){2-5} \cmidrule(lr){6-9}
Outcome & 
\scriptsize{BCAUSS} &
\scriptsize{DRAGON} &
\scriptsize{TARNET} &
\scriptsize{GANITE} &
\scriptsize{LinearReg.} &
\scriptsize{Rand.Forest} &
\scriptsize{XGBoost} &
\scriptsize{LightGBM} \\
\midrule
Metric 4 &
68\% &
45\% &
62\% &
15\% &
17\% &
14\% &
12\% &
14\% \\
Metric 5 &
58\% &
48\% &
64\% &
16\% &
14\% &
13\% &
16\% &
12\% \\
\bottomrule
\end{tabular}
}
\end{table*}

Advertising partners are the agencies and tool providers that have expertise in interacting with Ads products and help sellers/vendors in setting ad campaigns. Our team helps in identifying the actions that would enable partners to create maximum value for sellers/vendors. Such actions are considered opportunity for partners. Their impacts are measured on a wide list of business outcome metrics such as revenue and adspend.
Traditionally, it used to take 1-3 weeks of an Economist time to update the studies on ad-hoc requests. Since January 2023, we have been using OpportunityFinder to refresh the studies. Each refresh completes in a day with minimal human involvement.

OPF chooses GSC due to: number of total events $<$ 500,000, the number of treated units per monthly cohort $<$ 50 and control units $<$ 5,000.
In Table \ref{tab:hvo}, we show two such opportunities: adoption of \emph{X}, and adoption of \emph{Y}\footnote{Opportunity names and business metrics (see Tables~\ref{tab:hvo}, \ref{tab:vap}) are masked due to customer data policy.} by the partner for at-least one of their customers. Their lifts on three business metrics.
Comparing to results of prior studies, we can see delta between past and current downstream impact, which is caused due to behavioral and market changes over time. We further compare GSC results with DML and DNN models. While DML models return lift scores about 5\% to 10\% higher than GSC, DNN estimated lifts are from 20\% to hundred percent higher which are beyond acceptable range. ML-based models may over-estimate when input data is small.



\subsection{Opportunity for Advertisers}
This study estimates the effect of advertising partners on sellers/vendors outcomes related to Ads business, e.g., ad spend.
This study has been traditionally taking multi-weeks of scientist's effort for each refresh. In 2023, the study was expanded in both number of outcome variables and numbers of groups of partners and advertisers. Each combination of outcome and entity group is a separate causal study. OpportunityFinder helped accelerate the study so that all experiments were completed within a month.

With a large number of advertisers,
input data is redirected to DML and DNN causal models. Input panel data is then transformed into cohorts, before feature engineering and model training. Based on ATE and standard error results on validation datasets, OPF chooses \texttt{Rand.Forest+LinearDML} as final model. Our results were reviewed by domain experts and in range of results from prior studies. In Table~\ref{tab:vap}, we report lift metric returned by all possible models on a dataset. Three DNN models over-estimate treatment effect, and only GANITE yields numbers close to DML models.

\section{Conclusions and Future Work}
This paper presents OpportunityFinder (OPF), a codeless framework for causal inference studies, with a focus on panel data with binary treatment. Our experiments on multiple public, synthetic and internal datasets show that OPF can handle a diverse set of scenarios and our decision criteria for algorithm selection works well for given use-cases. We also see that in most of the cases, simpler algorithms like DML and GSC work well. We are able to use OPF on datasets ranging from small panel data to a large data with more than one million observations.

We are actively taking feature requests from current OPF users.
With causal discovery component, we will explore how hypothesis formulation before estimation can improve the estimation capability, especially with large set of observational data that a non expert user tends to provide.
We also aim to provide a master list of variables that can be collected for causal inference studies within our organization, and let OPF auto-shortlist covariates using data driven approaches for removing bias.
We plan to extend OPF by incorporating more estimators like meta learners, implement individual and heterogenous treatment effects, and support categorical and continuous treatments. 
With more and more causal inference algorithms being integrated into OPF, we will implement additional model selection, e.g., prediction/regression accuracy of base learners. Moreover, we will experiment model ensembles to provide the final output. Last but not least, we have refactored OpportunityFinder source code to make it a stand-alone library independent of AWS ecosystem.

\bibliographystyle{ACM-Reference-Format}
\bibliography{references}


\begin{thebibliography}{27}


\ifx \showCODEN    \undefined \def \showCODEN     #1{\unskip}     \fi
\ifx \showDOI      \undefined \def \showDOI       #1{#1}\fi
\ifx \showISBNx    \undefined \def \showISBNx     #1{\unskip}     \fi
\ifx \showISBNxiii \undefined \def \showISBNxiii  #1{\unskip}     \fi
\ifx \showISSN     \undefined \def \showISSN      #1{\unskip}     \fi
\ifx \showLCCN     \undefined \def \showLCCN      #1{\unskip}     \fi
\ifx \shownote     \undefined \def \shownote      #1{#1}          \fi
\ifx \showarticletitle \undefined \def \showarticletitle #1{#1}   \fi
\ifx \showURL      \undefined \def \showURL       {\relax}        \fi
\providecommand\bibfield[2]{#2}
\providecommand\bibinfo[2]{#2}
\providecommand\natexlab[1]{#1}
\providecommand\showeprint[2][]{arXiv:#2}

\bibitem[ihd({[n.\,d.]})]%
        {ihdbbench}
 \bibinfo{year}{[n.\,d.]}\natexlab{}.
\newblock \bibinfo{title}{{Causal Inference on IHDP: Benchmark}}.
\newblock \bibinfo{howpublished}{\url{
  https://paperswithcode.com/sota/causal-inference-on-ihdp}}.
\newblock


\bibitem[cau({[n.\,d.]})]%
        {causalAssumptions}
 \bibinfo{year}{[n.\,d.]}\natexlab{}.
\newblock \bibinfo{title}{{No Free Lunch in Causal Inference}}.
\newblock
  \bibinfo{howpublished}{\url{https://p-hunermund.com/2018/06/09/no-free-lunch-in-causal-inference/}}.
\newblock


\bibitem[Abadie et~al\mbox{.}(2010)]%
        {doi:10.1198/jasa.2009.ap08746}
\bibfield{author}{\bibinfo{person}{Alberto Abadie}, \bibinfo{person}{Alexis
  Diamond}, {and} \bibinfo{person}{Jens Hainmueller}.}
  \bibinfo{year}{2010}\natexlab{}.
\newblock \showarticletitle{Synthetic Control Methods for Comparative Case
  Studies: Estimating the Effect of California’s Tobacco Control Program}.
\newblock \bibinfo{journal}{\emph{J. Amer. Statist. Assoc.}}
  \bibinfo{volume}{105}, \bibinfo{number}{490} (\bibinfo{year}{2010}),
  \bibinfo{pages}{493--505}.
\newblock
\urldef\tempurl%
\url{https://doi.org/10.1198/jasa.2009.ap08746}
\showDOI{\tempurl}
\showeprint{https://doi.org/10.1198/jasa.2009.ap08746}


\bibitem[Abadie and Gardeazabal(2003)]%
        {Abadie2003}
\bibfield{author}{\bibinfo{person}{Alberto Abadie} {and}
  \bibinfo{person}{Javier Gardeazabal}.} \bibinfo{year}{2003}\natexlab{}.
\newblock \showarticletitle{The economic costs of conflict: A case study of the
  Basque Country}.
\newblock \bibinfo{journal}{\emph{American economic review}}
  (\bibinfo{year}{2003}), \bibinfo{pages}{113--132}.
\newblock


\bibitem[Angrist and Pischke(2009)]%
        {Angrist2009}
\bibfield{author}{\bibinfo{person}{Joshua~D. Angrist} {and}
  \bibinfo{person}{Jorn-Steffen Pischke}.} \bibinfo{year}{2009}\natexlab{}.
\newblock \bibinfo{booktitle}{\emph{Mostly harmless econometrics: An
  empiricist's companion}}.
\newblock \bibinfo{publisher}{Princeton university press}.
\newblock


\bibitem[Arkhangelsky et~al\mbox{.}(2021)]%
        {arkhangelsky2021synthetic}
\bibfield{author}{\bibinfo{person}{Dmitry Arkhangelsky}, \bibinfo{person}{Susan
  Athey}, \bibinfo{person}{David~A. Hirshberg}, \bibinfo{person}{Guido~W.
  Imbens}, {and} \bibinfo{person}{Stefan Wager}.}
  \bibinfo{year}{2021}\natexlab{}.
\newblock \bibinfo{title}{Synthetic Difference in Differences}.
\newblock
\newblock
\showeprint[arxiv]{1812.09970}~[stat.ME]


\bibitem[Bach et~al\mbox{.}(2022)]%
        {DoubleML2022Python}
\bibfield{author}{\bibinfo{person}{Philipp Bach}, \bibinfo{person}{Victor
  Chernozhukov}, \bibinfo{person}{Malte~S. Kurz}, {and} \bibinfo{person}{Martin
  Spindler}.} \bibinfo{year}{2022}\natexlab{}.
\newblock \showarticletitle{{DoubleML} -- {A}n Object-Oriented Implementation
  of Double Machine Learning in {P}ython}.
\newblock \bibinfo{journal}{\emph{Journal of Machine Learning Research}}
  \bibinfo{volume}{23}, \bibinfo{number}{53} (\bibinfo{year}{2022}),
  \bibinfo{pages}{1--6}.
\newblock
\urldef\tempurl%
\url{http://jmlr.org/papers/v23/21-0862.html}
\showURL{%
\tempurl}


\bibitem[Battocchi et~al\mbox{.}(2019)]%
        {econml}
\bibfield{author}{\bibinfo{person}{Keith Battocchi}, \bibinfo{person}{Eleanor
  Dillon}, \bibinfo{person}{Maggie Hei}, \bibinfo{person}{Greg Lewis},
  \bibinfo{person}{Paul Oka}, \bibinfo{person}{Miruna Oprescu}, {and}
  \bibinfo{person}{Vasilis Syrgkanis}.} \bibinfo{year}{2019}\natexlab{}.
\newblock \bibinfo{title}{{EconML}: {A Python Package for ML-Based
  Heterogeneous Treatment Effects Estimation}}.
\newblock \bibinfo{howpublished}{https://github.com/py-why/EconML}.
\newblock
\newblock
\shownote{Version 0.x}.


\bibitem[Card and Krueger(1994)]%
        {card1994minimum}
\bibfield{author}{\bibinfo{person}{David Card} {and} \bibinfo{person}{Alan~B
  Krueger}.} \bibinfo{year}{1994}\natexlab{}.
\newblock \showarticletitle{Minimum wages and employment: A case study of the
  fast-food industry in New Jersey and Pennsylvania}.
\newblock \bibinfo{journal}{\emph{The American economic review}}
  \bibinfo{volume}{84}, \bibinfo{number}{4} (\bibinfo{year}{1994}),
  \bibinfo{pages}{772--793}.
\newblock


\bibitem[Chen et~al\mbox{.}(2020)]%
        {chen2020causalml}
\bibfield{author}{\bibinfo{person}{Huigang Chen}, \bibinfo{person}{Totte
  Harinen}, \bibinfo{person}{Jeong-Yoon Lee}, \bibinfo{person}{Mike Yung},
  {and} \bibinfo{person}{Zhenyu Zhao}.} \bibinfo{year}{2020}\natexlab{}.
\newblock \bibinfo{title}{CausalML: Python Package for Causal Machine
  Learning}.
\newblock
\newblock
\showeprint[arxiv]{2002.11631}~[cs.CY]


\bibitem[Cheng and Hoekstra(2012)]%
        {NBERw18134}
\bibfield{author}{\bibinfo{person}{Cheng} {and} \bibinfo{person}{Mark
  Hoekstra}.} \bibinfo{year}{2012}\natexlab{}.
\newblock \bibinfo{booktitle}{\emph{Does Strengthening Self-Defense Law Deter
  Crime or Escalate Violence? Evidence from Castle Doctrine}}.
\newblock \bibinfo{type}{Working Paper} 18134. \bibinfo{institution}{National
  Bureau of Economic Research}.
\newblock
\urldef\tempurl%
\url{https://doi.org/10.3386/w18134}
\showDOI{\tempurl}


\bibitem[Chernozhukov et~al\mbox{.}(2018)]%
        {Chernozhukov2018}
\bibfield{author}{\bibinfo{person}{Victor Chernozhukov}, \bibinfo{person}{Denis
  Chetverikov}, \bibinfo{person}{Mert Demirer}, \bibinfo{person}{Esther Duflo},
  \bibinfo{person}{Christian Hansen}, \bibinfo{person}{Whitney Newey}, {and}
  \bibinfo{person}{James Robins}.} \bibinfo{year}{2018}\natexlab{}.
\newblock \showarticletitle{Double/debiased machine learning for treatment and
  structural parameters}.
\newblock \bibinfo{journal}{\emph{The Econometrics Journal}}
  \bibinfo{volume}{21}, \bibinfo{number}{1} (\bibinfo{year}{2018}),
  \bibinfo{pages}{C1--C68}.
\newblock


\bibitem[Dehejia and Wahba(2002)]%
        {10.1162/003465302317331982}
\bibfield{author}{\bibinfo{person}{Rajeev~H. Dehejia} {and}
  \bibinfo{person}{Sadek Wahba}.} \bibinfo{year}{2002}\natexlab{}.
\newblock \showarticletitle{{Propensity Score-Matching Methods for
  Nonexperimental Causal Studies}}.
\newblock \bibinfo{journal}{\emph{The Review of Economics and Statistics}}
  \bibinfo{volume}{84}, \bibinfo{number}{1} (\bibinfo{date}{02}
  \bibinfo{year}{2002}), \bibinfo{pages}{151--161}.
\newblock
\showISSN{0034-6535}
\urldef\tempurl%
\url{https://doi.org/10.1162/003465302317331982}
\showDOI{\tempurl}
\showeprint{https://direct.mit.edu/rest/article-pdf/84/1/151/1613304/003465302317331982.pdf}


\bibitem[Erickson et~al\mbox{.}(2020)]%
        {agtabular}
\bibfield{author}{\bibinfo{person}{Nick Erickson}, \bibinfo{person}{Jonas
  Mueller}, \bibinfo{person}{Alexander Shirkov}, \bibinfo{person}{Hang Zhang},
  \bibinfo{person}{Pedro Larroy}, \bibinfo{person}{Mu Li}, {and}
  \bibinfo{person}{Alexander Smola}.} \bibinfo{year}{2020}\natexlab{}.
\newblock \showarticletitle{{AutoGluon-Tabular: Robust and Accurate AutoML for
  Structured Data}}.
\newblock \bibinfo{journal}{\emph{arXiv preprint arXiv:2003.06505}}
  (\bibinfo{year}{2020}).
\newblock


\bibitem[Feurer et~al\mbox{.}(2020)]%
        {feurer-arxiv20a}
\bibfield{author}{\bibinfo{person}{Matthias Feurer}, \bibinfo{person}{Katharina
  Eggensperger}, \bibinfo{person}{Stefan Falkner}, \bibinfo{person}{Marius
  Lindauer}, {and} \bibinfo{person}{Frank Hutter}.}
  \bibinfo{year}{2020}\natexlab{}.
\newblock \showarticletitle{Auto-Sklearn 2.0: Hands-free AutoML via
  Meta-Learning}.
\newblock \bibinfo{journal}{\emph{arXiv:2007.04074 [cs.LG]}}
  (\bibinfo{year}{2020}).
\newblock


\bibitem[Flesch et~al\mbox{.}(2022)]%
        {autocausality}
\bibfield{author}{\bibinfo{person}{Timo Flesch}, \bibinfo{person}{Edward
  Zhang}, \bibinfo{person}{Guy Durant}, \bibinfo{person}{Mark~Harley Wen
  Hao~Kho}, {and} \bibinfo{person}{Egor Kraev}.}
  \bibinfo{year}{2022}\natexlab{}.
\newblock \bibinfo{title}{{Auto-Causality}: {A Python package for Automated
  Causal Inference model estimation and selection}}.
\newblock
  \bibinfo{howpublished}{https://github.com/transferwise/auto-causality}.
\newblock
\newblock
\shownote{Version 0.x}.


\bibitem[K{\"u}nzel et~al\mbox{.}(2019)]%
        {kunzel2019meta}
\bibfield{author}{\bibinfo{person}{S{\"o}ren~R K{\"u}nzel},
  \bibinfo{person}{Jasjeet~S Sekhon}, \bibinfo{person}{Peter~J Bickel}, {and}
  \bibinfo{person}{Bin Yu}.} \bibinfo{year}{2019}\natexlab{}.
\newblock \showarticletitle{Meta-learners for Estimating Heterogeneous
  Treatment Effects using Machine Learning}.
\newblock \bibinfo{journal}{\emph{Proceedings of the National Academy of
  Sciences}} \bibinfo{volume}{116}, \bibinfo{number}{10}
  (\bibinfo{year}{2019}), \bibinfo{pages}{4156--4165}.
\newblock


\bibitem[Lalonde(1986)]%
        {nsw}
\bibfield{author}{\bibinfo{person}{Robert Lalonde}.}
  \bibinfo{year}{1986}\natexlab{}.
\newblock \showarticletitle{Evaluating the Econometric Evaluations of Training
  Programs with Experiment Data}.
\newblock \bibinfo{journal}{\emph{American Economic Review}}
  \bibinfo{volume}{76} (\bibinfo{date}{02} \bibinfo{year}{1986}),
  \bibinfo{pages}{604--20}.
\newblock


\bibitem[LeDell and Poirier(2020)]%
        {H2OAutoML20}
\bibfield{author}{\bibinfo{person}{Erin LeDell} {and}
  \bibinfo{person}{Sebastien Poirier}.} \bibinfo{year}{2020}\natexlab{}.
\newblock \showarticletitle{{H2O AutoML: Scalable Automatic Machine Learning}}.
\newblock \bibinfo{journal}{\emph{7th ICML Workshop on Automated Machine
  Learning (AutoML)}} (\bibinfo{date}{July} \bibinfo{year}{2020}).
\newblock
\urldef\tempurl%
\url{https://www.automl.org/wp-content/uploads/2020/07/AutoML_2020_paper_61.pdf}
\showURL{%
\tempurl}


\bibitem[Rosenbaum and Rubin(1983)]%
        {rosenbaum1983central}
\bibfield{author}{\bibinfo{person}{Paul~R Rosenbaum} {and}
  \bibinfo{person}{Donald~B Rubin}.} \bibinfo{year}{1983}\natexlab{}.
\newblock \showarticletitle{The central role of the propensity score in
  observational studies for causal effects}.
\newblock \bibinfo{journal}{\emph{Biometrika}} \bibinfo{volume}{70},
  \bibinfo{number}{1} (\bibinfo{year}{1983}), \bibinfo{pages}{41--55}.
\newblock


\bibitem[Shalit et~al\mbox{.}(2017)]%
        {Shalit2017}
\bibfield{author}{\bibinfo{person}{Uri Shalit}, \bibinfo{person}{Fredrik~D.
  Johansson}, {and} \bibinfo{person}{David Sontag}.}
  \bibinfo{year}{2017}\natexlab{}.
\newblock \showarticletitle{Estimating individual treatment effect:
  generalization bounds and algorithms}. In
  \bibinfo{booktitle}{\emph{Proceedings of the 34th International Conference on
  Machine Learning-Volume 70}}. JMLR. org, \bibinfo{pages}{3076--3085}.
\newblock


\bibitem[Sharma et~al\mbox{.}(2019)]%
        {dowhy}
\bibfield{author}{\bibinfo{person}{Amit Sharma}, \bibinfo{person}{Emre
  Kiciman}, {et~al\mbox{.}}} \bibinfo{year}{2019}\natexlab{}.
\newblock \bibinfo{title}{Do{W}hy: {A Python package for causal inference}}.
\newblock \bibinfo{howpublished}{https://github.com/microsoft/dowhy}.
\newblock


\bibitem[Shi et~al\mbox{.}(2019)]%
        {shi2019adapting}
\bibfield{author}{\bibinfo{person}{Claudia Shi}, \bibinfo{person}{David~M.
  Blei}, {and} \bibinfo{person}{Victor Veitch}.}
  \bibinfo{year}{2019}\natexlab{}.
\newblock \bibinfo{title}{Adapting Neural Networks for the Estimation of
  Treatment Effects}.
\newblock
\newblock
\showeprint[arxiv]{1906.02120}~[stat.ML]


\bibitem[Tesei et~al\mbox{.}(2023)]%
        {tesei2023learning}
\bibfield{author}{\bibinfo{person}{Gino Tesei}, \bibinfo{person}{Stefanos
  Giampanis}, \bibinfo{person}{Jingpu Shi}, {and} \bibinfo{person}{Beau
  Norgeot}.} \bibinfo{year}{2023}\natexlab{}.
\newblock \showarticletitle{Learning end-to-end patient representations through
  self-supervised covariate balancing for causal treatment effect estimation}.
\newblock \bibinfo{journal}{\emph{Journal of Biomedical Informatics}}
  \bibinfo{volume}{140} (\bibinfo{year}{2023}), \bibinfo{pages}{104339}.
\newblock


\bibitem[Wager and Athey(2018)]%
        {wager2018estimation}
\bibfield{author}{\bibinfo{person}{Stefan Wager} {and} \bibinfo{person}{Susan
  Athey}.} \bibinfo{year}{2018}\natexlab{}.
\newblock \showarticletitle{Estimation and Inference of Heterogeneous Treatment
  Effects using Random Forests}.
\newblock \bibinfo{journal}{\emph{J. Amer. Statist. Assoc.}}
  \bibinfo{volume}{113}, \bibinfo{number}{523} (\bibinfo{year}{2018}),
  \bibinfo{pages}{1228--1242}.
\newblock


\bibitem[Xu(2017)]%
        {xu_2017}
\bibfield{author}{\bibinfo{person}{Yiqing Xu}.}
  \bibinfo{year}{2017}\natexlab{}.
\newblock \showarticletitle{Generalized Synthetic Control Method: Causal
  Inference with Interactive Fixed Effects Models}.
\newblock \bibinfo{journal}{\emph{Political Analysis}} \bibinfo{volume}{25},
  \bibinfo{number}{1} (\bibinfo{year}{2017}), \bibinfo{pages}{57–76}.
\newblock
\urldef\tempurl%
\url{https://doi.org/10.1017/pan.2016.2}
\showDOI{\tempurl}


\bibitem[Yoon et~al\mbox{.}(2018)]%
        {yoon2018ganite}
\bibfield{author}{\bibinfo{person}{Jinsung Yoon}, \bibinfo{person}{James
  Jordon}, {and} \bibinfo{person}{Mihaela van~der Schaar}.}
  \bibinfo{year}{2018}\natexlab{}.
\newblock \showarticletitle{{GANITE}: Estimation of Individualized Treatment
  Effects using Generative Adversarial Nets}. In
  \bibinfo{booktitle}{\emph{International Conference on Learning
  Representations}}.
\newblock
\urldef\tempurl%
\url{https://openreview.net/forum?id=ByKWUeWA-}
\showURL{%
\tempurl}


\end{thebibliography}


\section*{Appendix}
\appendix

\section{Overview of Causal Inference Models}
\label{app:models}
The following models are considered for the auto causal framework:

\begin{itemize}
\item \textbf{Synthetic Control (SC)} and \textbf{Generalized Synthetic Control (GSC)}: SC allows for comparative case studies using a weighted combination of control units to create a synthetic control unit. GSC extends this by considering interactive fixed effects models. The key assumption is that the outcome of treated units is a linear function of the outcomes of the control units in the absence of treatment. GSC allows relationship to vary over time, unlike traditional SC methods. Both methods are well suited for panel data with small sample sizes but require domain knowledge for selection of control units. The limitation for implementing GSC in the auto causal framework is computational inefficiency with large observational data or 
with more number of covariates. \cite{Abadie2003, xu_2017}

\item \textbf{Double Machine Learning (DML)}: DML leverages machine learning to estimate treatment effects in a semi-parametric manner, allowing for complex relationships. The key requirement for DML to work well is the availability of high-quality and diverse covariate data. DML can handle large datasets and does not specifically require panel data. It allows for different ML models to be used in the two stages, providing versatility. \cite{Chernozhukov2018}

\item \textbf{Causal Forests}: Causal Forests extend random forests to estimate heterogeneous treatment effects, offering flexibility and the ability to capture complex relationships. The key assumption is the unconfoundedness or ignorability assumption. It is not inherently designed for panel data and requires a relatively large sample size. The limitation for auto causal framework is that it does not handle panel data well and we did not find it to work 
well in our experiments. \cite{wager2018estimation}

\item \textbf{Neural Network based approaches}: Several approaches utilize neural networks for causal inference, each with its unique proposition: BCAUSS, Dragonnet and TARNet model treatment assignments and potential outcomes in a multi-task learning setup, allowing finding of least dissimilar treated and untreated 
observations. GANITE leverages the power of generative adversarial networks (GANs) to estimate individual treatment effects. They require relatively large and high-quality datasets, otherwise can over/under estimate the treatment effects. These methods can handle large datasets but are not specifically designed for panel data. \cite{tesei2023learning, shi2019adapting, Shalit2017, yoon2018ganite}

\item \textbf{Meta Learners}: Meta Learners apply machine learning methods to estimate treatment effects, offering the flexibility of using various base learners. The key assumption is that the base learners are correctly specified. They are not specifically designed for panel data and require a relatively large sample size. The limitation for auto causal framework is in the choice of base learner. \cite{kunzel2019meta}

\item \textbf{Difference in Differences (DiD)}: DiD compares the average change in outcome over time that occurs in the treatment group to the average change over time that happens in the control group. It's designed to handle unobserved, time-invariant confounders. It's a simple and intuitive method for panel data, widely used in economic studies. Problem 
with DiD is that is relies on strong parallel trends assumption that is often violated in the real world setting. We do not use DiD or variant Synthetic DiD in our implementations. \cite{card1994minimum, arkhangelsky2021synthetic}

\item \textbf{Propensity Score Matching (PSM)}: The propensity score is the conditional probability of receiving treatment given pre-treatment characteristics. This approach has been traditionally popular because of its simplicity, interpretability and ability to handle large covariates.  But DML is based on similar principle and overcomes the limitations that PSM has and is more robust. Complementing DML, SC methods do not require unconfoundedness assumption that PSM does. Therefore, we do not use PSM in the OPF. \cite{rosenbaum1983central} 

\end{itemize}


\section{Validation Tests for Treatment Effect Results}
\label{test:val}

To demonstrate the validation test for treatment effect results, we report refutation test outputs for LinearReg+LinearDML model and its ATE on Synthetic\#1 data in Table~\ref{tab:refute}. Sensitivity test for GSC model are reported in Table~\ref{tab:refute-gsc}. All refutation tests passed, placebo ATE are close to zero while other test ATE close to original model. This confirms the model is robust against changes in settings and estimated ATE is consistent.

\begin{table*}[t]
\centering
\caption{Refutation tests for DML model on Synthetic\#1 and \#2 sets}
\label{tab:refute}
{\normalsize
\begin{tabular}{lrlrl}
\toprule
 & Synth\#1 ATE &  & Synth\#2 ATE & \\
\midrule
Ground-truth    & 10.00 & & 20.00  \\
Model ATE    & 9.93 & & 19.04 \\
\midrule
Placebo test    & -0.08 & passed & -0.05 & passed \\
Random common cause test    & 9.98 & passed & 19.08 & passed  \\
Unobserved common cause test    & 9.98 & passed & 15.48 & passed  \\
Data-subset test    & 9.99 & passed & 19.00 & passed \\
\bottomrule
\end{tabular}%
}
\end{table*}

\begin{table*}[t]
\centering
\caption{Sensitivity tests for GSC model on partner data with Metric 2. The numbers represent \% uplift. All lifts are statistically significant}
\label{tab:refute-gsc}
{\normalsize
\begin{tabular}{lrlrl}
\toprule
 & Opportunity \emph{X} &  & Opportunity \emph{Y} & \\
\midrule
Overall model    & 7.8\% & & 12.0\%  \\
\midrule
Remove covariates test    & 14.3\% & passed & 34.5\% & passed \\
Random downsample test    & 7.0\% & passed & 11.8\% & passed  \\
Reduced period for SC weights test    & 7.8\%  & passed & 11.2\% & passed  \\
\bottomrule
\end{tabular}%
}
\end{table*}


\begin{figure}[t]
    \centering
    \begin{subfigure}{0.45\textwidth}
        \centering
        \includegraphics[width=\linewidth]{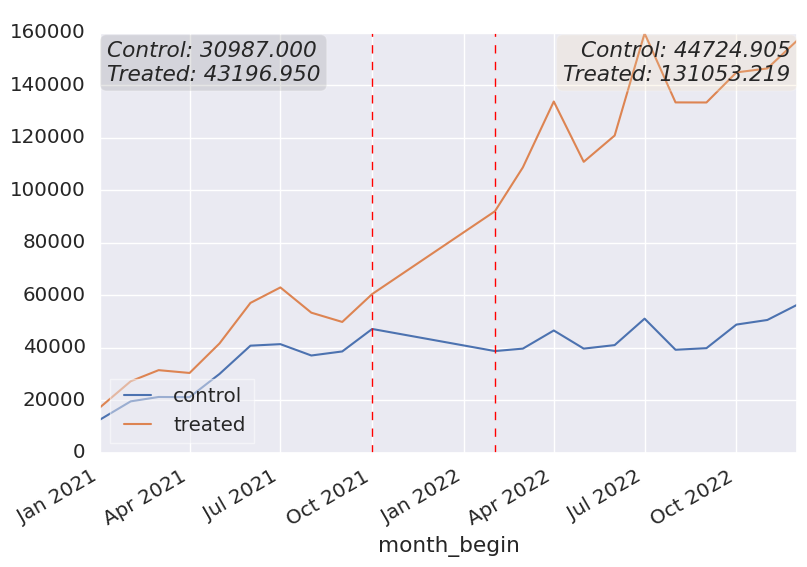}
        \caption{Metric 4}
        \label{fig:subfig-vap-a}
    \end{subfigure}
    \hfill
    \begin{subfigure}{0.45\textwidth}
        \centering
        \includegraphics[width=\linewidth]{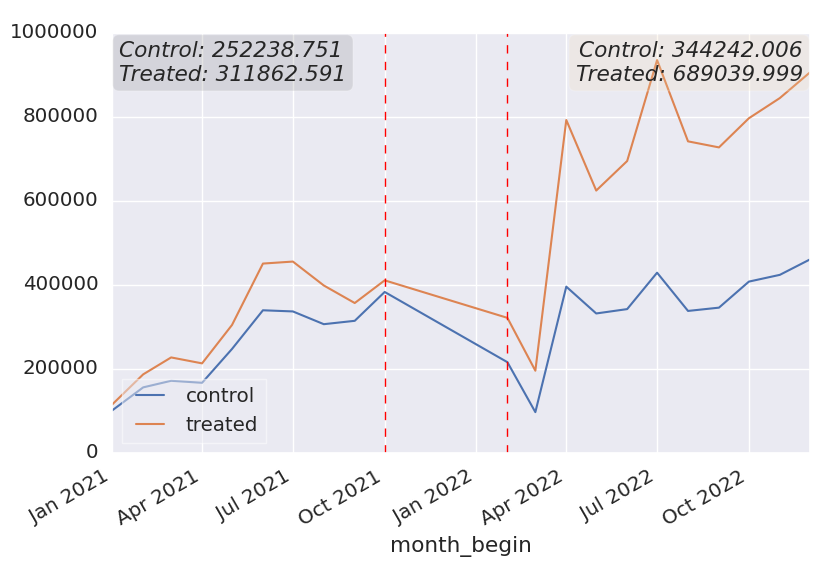}
        \caption{Metric 5}
        \label{fig:subfig-vap-b}
    \end{subfigure}
    \caption{Average outcome metric of treated vs. control units over time in advertiser data.}
    \label{fig:vap}
\end{figure}

\section{Visualization of Outcome Variables in Different Data}
\label{app:vis}

\begin{figure*}[t]
\includegraphics[scale=0.4]{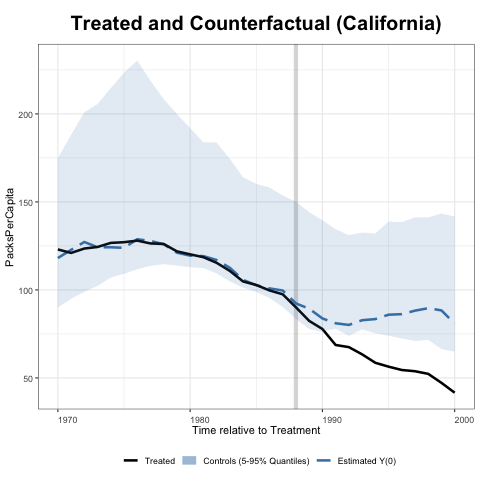}
\centering
\caption{Synthetic control fit on smoking data without covariate. The pre-treatment fit 
is good.}
\label{fig:gsc_smoking_nocovar}
\end{figure*}

\begin{figure*}[t]
    \centering
    \begin{subfigure}{0.45\textwidth}
        \centering
        \includegraphics[width=\linewidth]{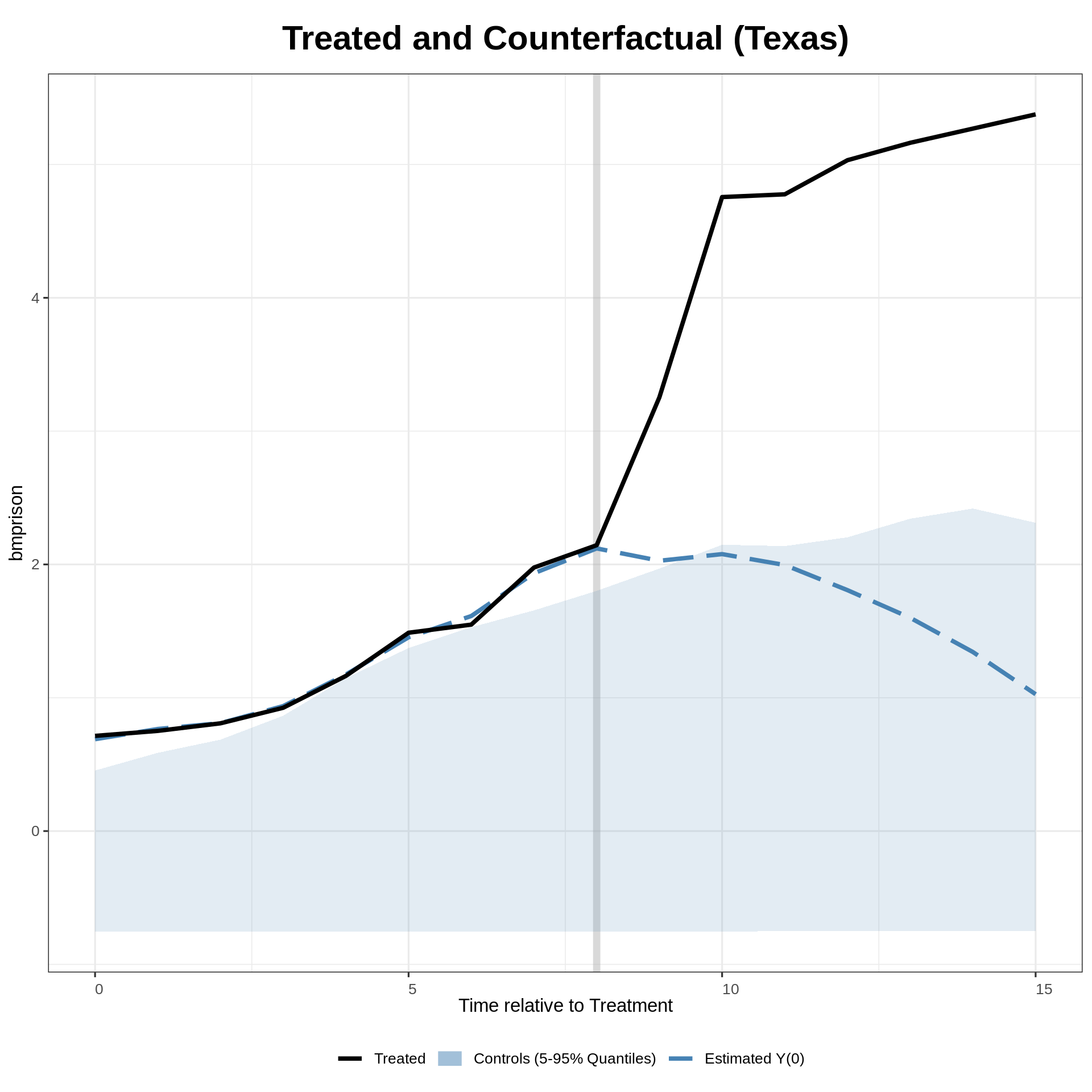}
        \caption{Black male incarceration}
        \label{fig:subfig-a}
    \end{subfigure}
    \hfill
    \begin{subfigure}{0.45\textwidth}
        \centering
        \includegraphics[width=\linewidth]{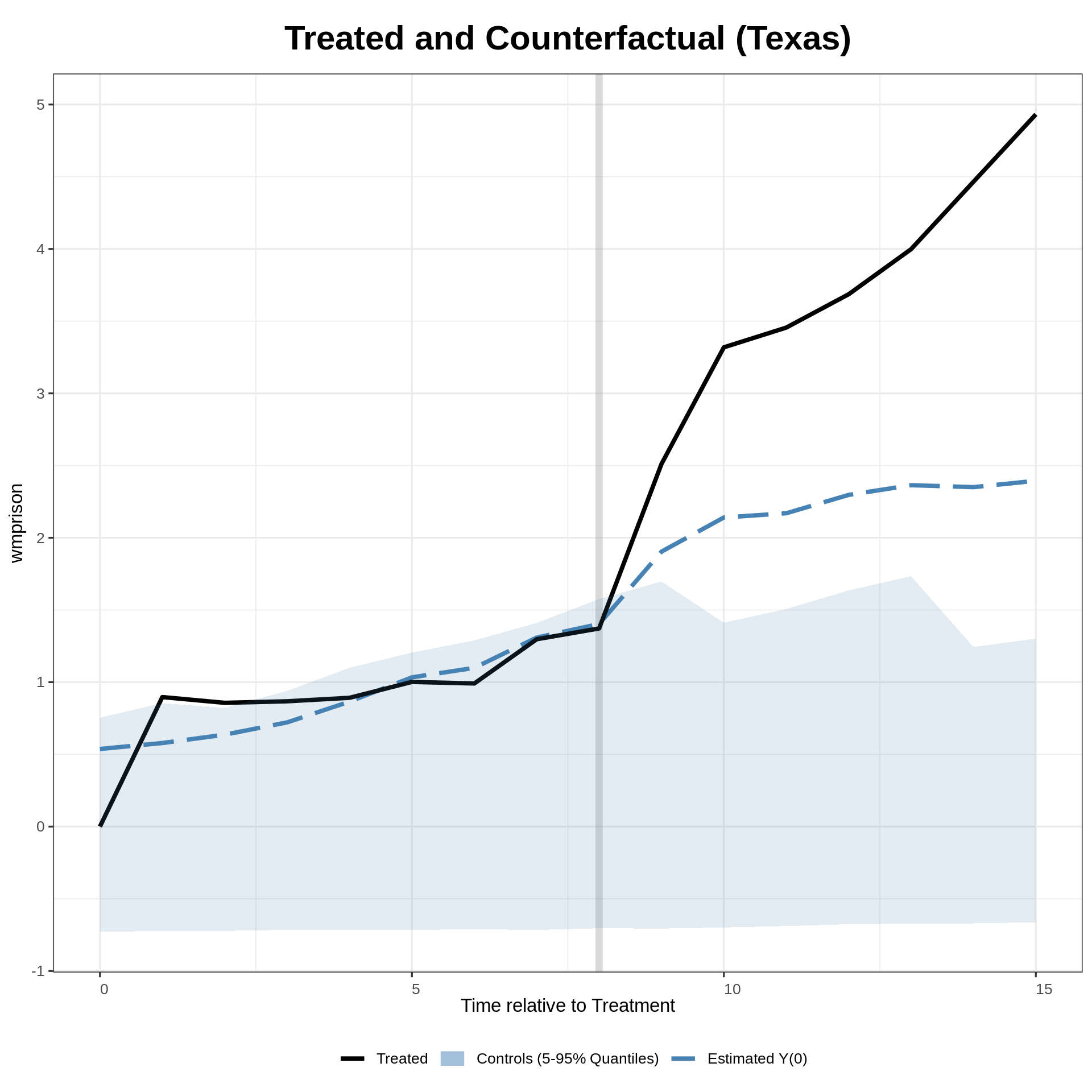}
        \caption{White male incarceration}
        \label{fig:subfig-b}
    \end{subfigure}
    
    \vspace{10pt} 

    \begin{subfigure}{0.45\textwidth}
        \centering
        \includegraphics[width=\linewidth]{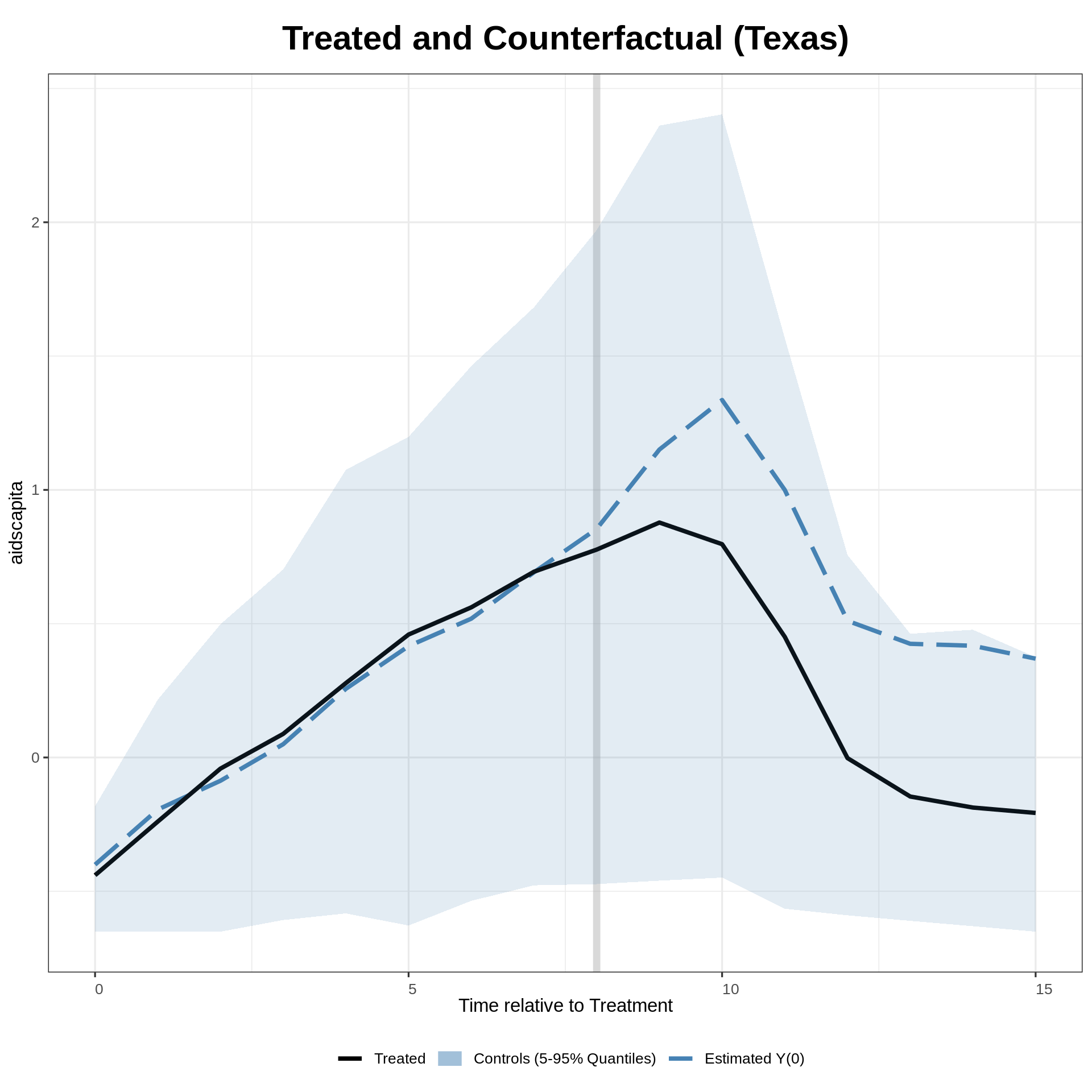}
        \caption{Aids per capita}
        \label{fig:subfig-c}
    \end{subfigure}
    \hfill
    \begin{subfigure}{0.45\textwidth}
        \centering
        \includegraphics[width=\linewidth]{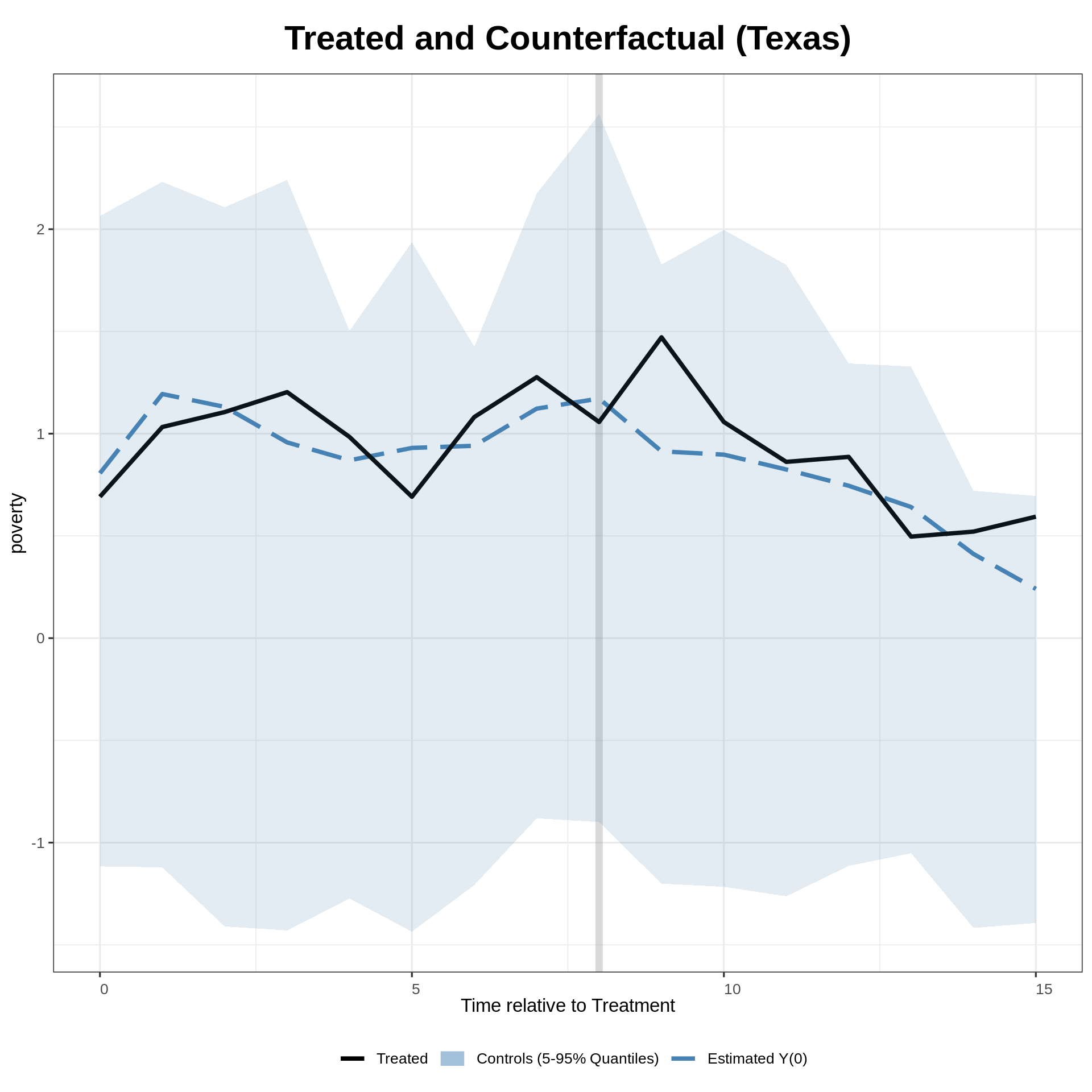}
        \caption{Poverty}
        \label{fig:subfig-d}
    \end{subfigure}
    \caption{Pre treatment synthetic control fit and post treatment diversion of different metrics on Texas data.}
    \label{fig:texas}
\end{figure*}

We display data plots generated by OpportunityFinder when running different datasets. Figure~\ref{fig:vap} plots average outcome of treated (orange line) and control (blue line) units for a cohort. Dash-red vertical bars indicate start and end date of cohort. These plots are generated by our data processing module. For Smoking and Texas data, Figures \ref{fig:gsc_smoking_nocovar}, \ref{fig:texas} are generated by our GSC model that black line shows time-series of outcome values of treated unit, and dash-blue line show that of synthesized control.


\section{Additional Results}

\subsection{\textit{Texas} data}

Table~\ref{tab:texas} shows impact on black and white male incarceration from prison expansion in Texas since 1993. The numbers represent average percentage lift on the respective observational metric in Texas vs. other states due to the expansion. The covariates used to remove bias include poverty rates, white male incarceration, percentage of population between 15 and 19, income, unemployment rate and AIDS mortality. We see that the Texas had an average of 2x (100\%) more black male incarceration compared to the other states after they started prison expansion in 1993 till 2000. The increase was low but non zero (38\%) for white male incarceration during the same period in Texas.

\subsection{\textit{NSW} and \textit{Castle} data}

\begin{table*}[t]
\centering
\caption{Average percentage lift on black and white male incarceration from prison expansion in Texas since 1993.}
\label{tab:texas}
{\normalsize
\begin{tabular}{llll}
\toprule
black-male prison & white-male prison \\
\midrule
100\%    & 38\%  \\
\bottomrule
\end{tabular}%
}
\end{table*}

\begin{table*}[t]
\centering
\caption{Comparison of \% lift on Castle datasets using DML models from OPF vs. previous research works}
\label{tab:castle}
{\normalsize
\begin{tabular}{r|rrrr}
\toprule
\multicolumn{1}{c|}{Previous Research} & \multicolumn{4}{c}{+LinearDML} \\
\midrule
Cheng'12 \cite{NBERw18134} &
LinearReg. &
Rand.Forest &
XGBoost &
LightGBM \\
\midrule
8\% &
\textbf{7\%} &
4\% &
4\% &
50\% \\
\bottomrule
\end{tabular}%
}
\end{table*}

\begin{table*}[t]
\centering
\caption{Comparison of ATE on NSW datasets using DML models from OPF vs. previous research works}
\label{tab:nsw}
{\normalsize
\begin{tabular}{rr|rrrr}
\toprule
\multicolumn{2}{c|}{Previous Research} & \multicolumn{4}{c}{+LinearDML} \\
\midrule
Lalonde'86 \cite{nsw} &
Dehejia'02 \cite{10.1162/003465302317331982} &
LinearReg. &
Rand.Forest &
XGBoost &
LightGBM \\
\midrule
900 &
1300-1800 &
286 &
\textbf{776} &
728 &
637 \\
\bottomrule
\end{tabular}%
}
\end{table*}

In this section, we show results on two additional datasets, NSW \cite{nsw} 
and Castle \cite{NBERw18134}. Caslte is a panel data with year level 
information for 10 years, covering 50 states out of which 21 adopted the  
castle doctrine law. Castle law designates a person's abode or any legally occupied place (for example, a vehicle or home) as a place in which that person has protections and immunity permitting, in certain circumstances, to use force (up to and including deadly force) to defend oneself against an intruder, free from legal prosecution for the consequences of the force used. The study done by \cite{NBERw18134} aimed at finding the effect of castle doctrine law on increase in homicide. This data has 550 rows and 170 features (potential covariates), and based on researcher's outcome, expected lift in homicide of 8\% (we are accessing if OPF models reproduce the study). We see that in such small data sizes with large number of potential covariates, linear models do the best, and boosted trees can be very off. 

NSW is a famous experimental data that is complemented with additional synthetic data where researchers added selection bias in the control population. This is used in multiple research works to replicate the results of randomized trials. It contains ~50,000 rows and 180 features, and has 100 simulated variations. It is not a panel data and we tested it with DML variations. We observe underestimation compared to 
other research works but closest results from Random Forest based model. 

\end{document}